# Adaptive Reliability Analysis for Multi-fidelity Models using a Collective Learning Strategy

Chi Zhang[1], Chaolin Song[1,2], Abdollah Shafieezadeh[1]

[1] Risk Assessment and Management of Structural and Infrastructure Systems (RAMSIS) Lab, Department of Civil, Environmental, and Geodetic Engineering, The Ohio State University, Columbus, OH, 43210, United States

[2] Department of Bridge Engineering, Tongji University, Shanghai, 200092, China

**ABSTRACT**

In many fields of science and engineering, models with different fidelities are available. Physical experiments or detailed simulations that accurately capture the behavior of the system are regarded as high-fidelity models with low model uncertainty, however, they are expensive to run. On the other hand, simplified physical experiments or numerical models are seen as low-fidelity models that are cheaper to evaluate. Although low-fidelity models are often not suitable for direct use in reliability analysis due to their low accuracy, they can offer information about the trend of the high-fidelity model thus providing the opportunity to explore the design space at a low cost. This study presents a new approach called adaptive multi-fidelity Gaussian process for reliability analysis (AMGPRA). Contrary to selecting training points and information sources in two separate stages as done in state-of-the-art mfEGRA method, the proposed approach finds the optimal training point and information source simultaneously using the novel collective learning function ($CLF$). $CLF$ is able to assess the global impact of a candidate training point from an information source and it accommodates any learning function that satisfies a certain profile. In this context, $CLF$ provides a new direction for quantifying the impact of new training points and can be easily extended with new learning functions to adapt to different reliability problems. The performance of the proposed method is demonstrated by three mathematical examples and one engineering problem concerning the wind reliability of transmission towers. It is shown that the proposed method achieves similar or higher accuracy with reduced computational costs compared to state-of-the-art single and multi-fidelity methods. A key application of AMGPRA is high-fidelity fragility modeling using complex and costly physics-based computational models.

## 1. Introduction

Reliability analysis in many applications is concerned with estimating the probability of a distinct event of interest, often the failure event, considering the uncertainties that affect the occurrence of the event. In this context, reliability is an essential performance measure for various systems. The failure probability $P_f$ can be calculated by solving the multifold probability integral defined as:

$$P_f = P(g(\boldsymbol{x}) \leq 0) = \int_{g(\boldsymbol{x}) \leq 0} f(\boldsymbol{x}) d\boldsymbol{x} \qquad (1)$$

where $\boldsymbol{x}$ is the vector of the random variables and $g(\boldsymbol{x})$ is the limit state function, which determines the status of the system: $g(\boldsymbol{x}) \leq 0$ indicates failure and $g(\boldsymbol{x}) > 0$ otherwise. Moreover, $f(\boldsymbol{x})$ is the joint probability density function of $\boldsymbol{x}$. Advances in science have afforded uncovering complexities in physical phenomena, which coupled with technological advancements in modeling and simulation have given rise to complex computational models. Analyses of the events of interest via limit state functions may implicitly include such high-fidelity simulations. Thus, it is often prohibitive to solve the above integral directly. Simulation methods, such as Monte Carlo Simulation (MCS) [1] and Importance Sampling (IS) [1], are alternative techniques. In these methods, a large number of realizations of computational models are needed to obtain an estimate of the failure probability. For example, in MCS the failure probability is estimated using the following equation:

$$P_f = \int f(\boldsymbol{x}) I(\boldsymbol{x}) d\boldsymbol{x} \approx \frac{1}{N_{MCS}} \sum_{i=1}^{N_{MCS}} I(\boldsymbol{x}_i) \qquad (2)$$

where $I(\boldsymbol{x})$ is the indicator function with the property that $I(\boldsymbol{x}) = 1$ when $g(\boldsymbol{x}) \leq 0$ and $I(\boldsymbol{x}) = 0$ otherwise. Moreover, $N_{MCS}$ is a sufficiently large population of MCS and $\boldsymbol{x}_i$ is the $i$th sample point in the MCS population. However, MCS often requires numerous model evaluations to ensure small coefficient of variation of failure probability. This requirement leads to extremely high computational costs. On the other hand, IS may reduce the computational via improved sampling. However, this method requires a suitable importance distribution that can be difficult to acquire. Approximation methods, such as first and second order reliability method (FORM and SORM) [2], [3], utilize Taylor series expansion of the limit state function to approximate the failure probability. These methods often have significantly lower computational demand compared to simulation-based methods. However, their results may suffer from significant approximation errors when dealing with highly nonlinear problems.

Accuracy and efficiency are often two conflicting objectives in reliability analysis. Metamodel-based methods have recently received much attention, as they can substantially reduce the otherwise very high computational costs of reliability analysis. In these methods, metamodels – also called surrogate models – are constructed with limited number of function evaluations to mimic the original time-consuming limit state functions and subsequently estimate the failure probability. If the surrogate models are well-constructed, this group of methods can be both efficient and accurate. Many metamodels have been used for reliability analysis, including Polynomial Response Surface [4]–[6], Polynomial Chaos Expansion (PCE) [7], Support Vector Regression (SVR) [8], [9], and Kriging [10]. Among them, Kriging metamodel, also referred to as Gaussian process (GP) metamodel, has gained significant popularity in the community of reliability analysis due to its ability to provide uncertainty information. The uncertainty information can help to identify the 'best' training points for adaptive refinement of the metamodel. Jones et al. [11] proposed Efficient Global Optimization to adaptively solve global optimization problems using Kriging models. Their work inspired Bichon et al. [12] to propose Efficient Global Reliability Analysis (EGRA) that used Kriging metamodel to perform reliability analysis. Echard et al. [10] developed an active learning reliability method that combines Kriging and Monte Carlo Simulation (AK-MCS). The popular $U$ learning function was proposed in that study to select best training points among candidate design points. Bect et al. [13] proposed stepwise uncertainty reduction strategies from a Bayesian formulation of the reliability analysis problem. Picheny et al. [14] proposed an adaptive strategy to construct a Kriging metamodel based on an explicit trade-off between reduction in global uncertainty and exploration of regions of interest. Gaspar et al. [15] proposed an adaptive Kriging-based trust region method to search for the design point with IS. Xiao et al. [16] used active learning Kriging to address system reliability based-design optimization problems. Zhang et al. [17] solved a value of information analysis problem with an adaptive Kriging approach. Wang and Shafieezadeh [18] developed analytical confidence intervals (CIs) for failure probability estimates using adaptive Kriging. The same authors used adaptive Kriging approach to efficiently solve reliability updating problems with equality information [19]. Zhang et al. [20] developed an error-based stopping criterion for adaptive Kriging-based reliability updating problems with equality information. Rahimi et al. [21] performed failure analysis of soil slopes using adaptive Kriging approach with an effective sampling region and an accurate stopping criterion.

The above methods are built on the premise that data for the construction of metamodels are available from a single fidelity source, often a high-fidelity model. High-fidelity models often entail physical experiments or detailed simulations that accurately capture the behavior of the engineering system. Evaluation of these high-fidelity models can be extremely expensive, thus limiting the number of runs that can be afforded. Although high-fidelity models per se produce low model uncertainty, the accuracy of the metamodels cannot be guaranteed with a limited amount of expensive high-fidelity data. In many scientific domains, models with multiple fidelities are available for analyzing the same phenomena of interest. For example, varying the resolution of a finite element model will yield models with different fidelities. While coarsely meshed models may not be as accurate as the fine-meshed counterpart, they can be significantly faster and still offer useful information about the phenomena, albeit less accurately. Multi-fidelity metamodels have been proposed to leverage different information sources ranging from accurate but expensive high-fidelity data to less accurate but cheap low-fidelity data [22]–[25]. For instance, Kennedy and O'Hagan [22] suggested an autoregressive model that combines data from multiple fidelity levels. Qian et al. [23] proposed a two-step Bayesian approach to integrate results from high-fidelity and low-fidelity computer experiments. Subsequently, Qian and Wu [24] developed a hierarchical GP model for modeling and integrating high-fidelity and low-fidelity data. Le Gratiet [25] extended the scaling parameter in the multi-fidelity GP model to a more practical condition and derived the analytical expressions for the posteriors of model parameters. Li and Jia [26] proposed a general multi-fidelity GP model integrating low-fidelity data and high-fidelity data considering censoring in high-fidelity data. The application of multi-fidelity GP models can be found in design optimization problems as well [27]–[31]. For instance, Forrester et al. [28] applied correlated GP-based approximations to optimization where multiple levels of analysis are available, using an extension to the geostatistical method of co-kriging. Poloczek et al. [30] proposed a Bayesian method for multi-information source optimization (MISO) to optimize an expensive-to-evaluate black-box objective function, while utilizing cheaper but biased and noisy approximations. Zhou et al. [31] developed a multi-fidelity metamodel assisted robust optimization approach that quantifies and takes into consideration the interpolation uncertainty of the MF metamodel and design variable uncertainty.

In the context of reliability analysis using active learning metamodels, the application of multi-fidelity GP methods is scarce. As a matter of fact, multi-fidelity metamodels greatly fit the adaptive reliability analysis paradigm and can be applied to boost the efficiency of failure probability analyses. In single-fidelity adaptive reliability analysis, the domain of interest is gradually explored using training points from a single information source, often a high-fidelity model. However, the accuracy of the metamodel is of importance only in the vicinity of the limit state, and other regions do not require high accuracy for exploration. On the other hand, the low-fidelity models are not suitable for direct use in reliability analysis due to their low accuracy, however, they can be used to explore regions other than the vicinity of the currently known limit state at a significantly lower cost. Chaudhuri et al. [32] developed multi-fidelity EGRA (mfEGRA) method, which is a multi-fidelity extension of the popular EGRA method to solve reliability analysis problems with multi-fidelity data. In each iteration of their method, a training point is firstly located using Expected Feasible Function ($EFF$) in EGRA for the current high-fidelity metamodel, then the information source is selected using weighted lookahead information gain. The metamodel is adaptively refined by adding the selected training point in an information source until a stopping criterion based on $EFF$ is satisfied. At the end, the constructed metamodel is used to perform reliability analysis using a simulation method, i.e., MCS. In mfEGRA, the selection of next training point and the selection of information source are separate, which may lead to sub-optimal solutions. For example, a training point from a low-fidelity information source may be more cost-effective than the point located by $EFF$ for the high-fidelity metamodel. A selection process that chooses the training point and information source simultaneously can potentially improve the construction of the metamodel. Yi et al. [33] proposed an active-learning method based on the multi-fidelity Kriging model for structural reliability analysis (AMK-MCS+AEFF). They proposed a learning function named AEFF that integrates $EFF$, a cross-correlation function and a sampling density function to choose the training point and information source together. The method was proven accurate and efficient through several examples. However, the framework is limited to bi-fidelity problems, i.e., the method can only deal with problems with two fidelities [33], while in reality multiple levels of fidelities can exist.

This paper proposes adaptive multi-fidelity Gaussian process reliability analysis (AMGPRA) to solve reliability problems with multi-fidelity data. A novel approach to selecting the best training points from a set of candidate design samples is proposed. Referred to as collective learning function ($CLF$), this method captures the degree to which any training point from any information source can improve the accuracy of reliability estimation by the metamodel. Compared with other learning functions that consider only the local impact of a new training point, $CLF$ considers the global impact to facilitate the refinement of the metamodel in a more efficient manner. In addition, $CLF$ is adaptable to any learning function that satisfies a certain profile at the core. The proposed method uses $CLF$ to find the optimal training points and information sources simultaneously, resulting in an optimal construction of the metamodel. To demonstrate the performance, the proposed method is applied to three numerical examples in addition to a practical engineering example that is concerned with the wind reliability of a real transmission tower with a low- and high-fidelity finite element model.

The rest of this paper is organized into six sections. Section 2 provides a review of adaptive GP reliability methods. Section 3 introduces the multi-fidelity GP used in the proposed method and provides a review of mfEGRA. The proposed method is introduced in detail in Section 4. The performance of the proposed method is demonstrated through several examples in Section 5. The conclusions are drawn in Section 6.

## 2. Adaptive Gaussian process reliability methods: a brief review

This section provides a brief overview of adaptive GP reliability analysis methods. These techniques have gained significant attention in reliability analysis field [10], [12], [34]–[36], due in part to their ability to estimate the uncertainty around their predictions. The predictions of GP metamodels follow normal distributions and the probabilistic characteristics of these normal distributions are also provided by the metamodel. That is, for an untried point $x^*$ the GP model yields the predicted value mean $\mu_{GP}(x^*)$ and its corresponding posterior variance $\sigma_{GP}^2(x^*)$. A larger posterior variance for a point indicates that the metamodel is more uncertain about the prediction at that point. The uncertainty information in the form of $\sigma_{GP}^2(x^*)$ is used in adaptive reliability analysis methods along with the mean $\mu_{GP}(x^*)$ to guide the selection of 'best' training points.

The main underlying procedure in adaptive Gaussian process reliability methods can be generalized as follows: (1) construct an initial GP metamodel with a set of randomly generated training points $X_{tr}$; (2) perform reliability analysis with the current metamodel; (3) check the stopping criterion. If satisfied, end the process and provide the output, if not, go to the next step; (4) use a learning function that takes advantage of the uncertainty information to locate the next 'best' training point and add the point to $X_{tr}$, then update the metamodel with the current $X_{tr}$ and go to Step (2). A flowchart of the process is shown in Fig. 1.

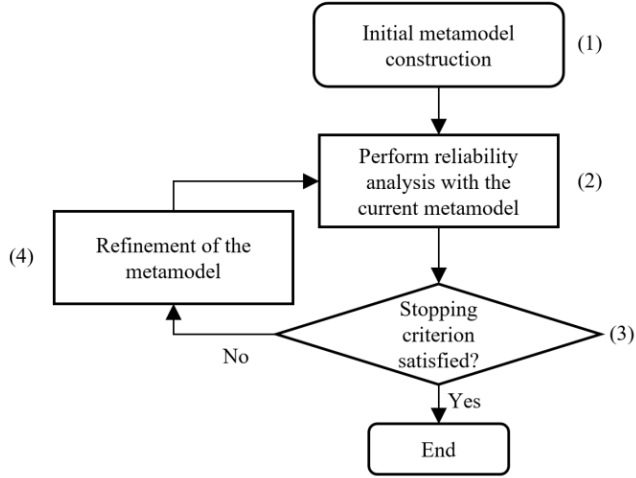

Fig.1 Flowchart of adaptive Gaussian process reliability methods

One of the most essential steps here is Step (2), where all the training points, except for the randomly generated initial set in Step (1), are determined. These training points can directly affect the performance of the adaptive method, as they determine the accuracy of the metamodel. Many studies in GP metamodel-based reliability analysis are focused on learning functions that use the uncertainty information to determine the 'best' training point. For instance, in EGRA, Bichon et al. [12] proposed Expected Learning Function ($EFF$) to locate the training points that are close to the limit state and have large uncertainty. The mathematical expression of $EFF$ is as follows:

$$\begin{aligned}
EFF(\pmb{x}^*) &= \int_{a-\epsilon(x)}^{a+\epsilon(x)} [\epsilon(\pmb{x}^*) - |a - h|]\phi(h; \mu_{GP}(\pmb{x}^*), \sigma_{GP}(\pmb{x}^*))dh \\
&= (\mu_{GP}(\pmb{x}^*) - a)\left[2\Phi\left(\frac{a - \mu_{GP}(\pmb{x}^*)}{\sigma_{GP}(\pmb{x}^*)}\right) - \Phi\left(\frac{a^- - \mu_{GP}(\pmb{x}^*)}{\sigma_{GP}(\pmb{x}^*)}\right) - \Phi\left(\frac{a^+ - \mu_{GP}(\pmb{x}^*)}{\sigma_{GP}(\pmb{x}^*)}\right)\right] \\
&\quad -\sigma_{GP}(\pmb{x}^*)\left[2\phi\left(\frac{a - \mu_{GP}(\pmb{x}^*)}{\sigma_{GP}(\pmb{x}^*)}\right) - \phi\left(\frac{a^- - \mu_{GP}(\pmb{x}^*)}{\sigma_{GP}(\pmb{x}^*)}\right) - \phi\left(\frac{a^+ - \mu_{GP}(\pmb{x}^*)}{\sigma_{GP}(\pmb{x}^*)}\right)\right] \\
&\quad + 2\sigma_{GP}(\pmb{x}^*)\left[\Phi\left(\frac{a^+ - \mu_{GP}(\pmb{x}^*)}{\sigma_{GP}(\pmb{x}^*)}\right) - \Phi\left(\frac{a^- - \mu_{GP}(\pmb{x}^*)}{\sigma_{GP}(\pmb{x}^*)}\right)\right]
\end{aligned} \quad (3)$$

where $\phi(\cdot)$ is the standard normal probability density function (PDF), $\Phi$ is the standard normal cumulative density function (CDF), $a = 0$, $\epsilon(\pmb{x}^*) = 2\sigma_{GP}(\pmb{x}^*)$, $a^+ = a + \epsilon(\pmb{x}^*)$ and $a^- = a - \epsilon(\pmb{x}^*)$. The term $[\epsilon(\pmb{x}^*) - |a - h|]$ measures the proximity of the untried point to the limit state $g(\pmb{x}^*) = a$, and is weighted by the term $\phi(h; \mu_{GP}(\pmb{x}^*), \sigma_{GP}(\pmb{x}^*))$. In AK-MCS, Echard et al. [10] proposed $U$ learning function, which has the following mathematical expression:

$$U(\pmb{x}^*) = \frac{|\mu_{GP}(\pmb{x}^*)|}{\sigma_{GP}(\pmb{x}^*)} \quad (4)$$

This learning function gives more weight to points close to the predicted limit state rather than those that are further away but with higher posterior variance like $EFF$. Most current studies on metamodel-based reliability analysis are focused on problems with only one high-fidelity model. To develop metamodels for multi-fidelity data, not only the 'best' training point should be identified, but also the 'best' information source should be chosen, leading to a more complex optimization problem. The discussion is provided in the next section.

## 3. Multi-fidelity Gaussian process reliability analysis

While multi-fidelity GP metamodels have been developed and applied in several studies as reviewed in Section 1, there are very few applications of this method in reliability analysis applications. In this section, the multi-fidelity GP metamodel is first introduced followed by a review of mfEGRA.

### 3.1 Multi-fidelity Gaussian process metamodel

The multi-fidelity GP metamodel introduced by Poloczek et al. [30] is adopted in this study. This metamodel fuses information from multiple information sources into a single GP metamodel. Assume that for a reliability analysis problem, $k+1$ models of the limit state function are available. Among them, there are one high-fidelity model $g_0$ and $k$ low-fidelity models $g_l, l = 1,2,...,k$. The multi-fidelity GP metamodel is constructed using data from all $k+1$ information sources and can provide predictions for all $k+1$ models at a location $\boldsymbol{x}$ simultaneously. Let $\hat{g}(l, \boldsymbol{x})$ denote the multi-fidelity GP metamodel. Assume a GP approximation for the high-fidelity model $g_0$ can be expressed as $\hat{g}(0, \boldsymbol{x}) \sim GP(\mu_0, \Sigma_0)$, where $\mu_0$ is the mean function for $\hat{g}(0, \boldsymbol{x})$ and $\Sigma_0$ is the covariance kernel for $\hat{g}(0, \boldsymbol{x})$. For each $l > 0$, suppose an independent GP approximation can be constructed for the model discrepancy between $g_l$ and $g_0$, which is denoted as $\delta_l \sim GP(\mu_l, \Sigma_l)$. Here, $\mu_l$ is the mean function for $\delta_l$ and $\Sigma_l$ is the covariance kernel for $\delta_l$. In this study, it is further assumed that $\mu_l(x) = 0, l = 1,2,...,k$ in the absence of a strong belief for the direction of all model discrepancies. One can also set $\mu_l(x)$ as a constant estimated with maximum likelihood estimation (MLE) if one has a strong belief that the model discrepancy is consistently positive or negative. By definition, the metamodels for low-fidelity data can be constructed as $\hat{g}(l, \boldsymbol{x}) = \hat{g}(0, \boldsymbol{x}) + \delta_l, l = 0,1,2,...,k$ with $\delta_0 = 0$.

With the above assumptions, given any finite collection of training points from $k+1$ information sources, $\hat{g}(l, \boldsymbol{x})$ is a GP and $\hat{g} \sim GP(\mu_{pr}, \Sigma_{pr})$. As $\hat{g}(l, \boldsymbol{x})$ is the sum of independent multivariate normal random vectors, it is itself a multivariate normal random vector. When $l = 0$, i.e., for the high-fidelity metamodel, the priors are:

$$\mu_{pr}(0, \boldsymbol{x}) = \mathbb{E}(\hat{g}(0, \boldsymbol{x})) = \mu_0(\boldsymbol{x})$$
$$\Sigma_{pr}((0, \boldsymbol{x}), (0, \boldsymbol{x}')) = Cov(\hat{g}(0, \boldsymbol{x}), \hat{g}(0, \boldsymbol{x}')) = \Sigma_0(\boldsymbol{x}, \boldsymbol{x}') \quad (5)$$

When $l = 1,2,...,k$, i.e., for the low-fidelity metamodels, the priors are:

$$\mu_{pr} = \mathbb{E}(\hat{g}(0, \boldsymbol{x})) + \mathbb{E}(\delta_l(\boldsymbol{x})) = \mu_0(\boldsymbol{x})$$
$$\Sigma_{pr}((l, \boldsymbol{x}), (l', \boldsymbol{x}')) = Cov(\hat{g}(0, \boldsymbol{x}) + \delta_l(\boldsymbol{x}), \hat{g}(0, \boldsymbol{x}') + \delta_{l'}(\boldsymbol{x}')) \quad (6)$$
$$= \Sigma_0(\boldsymbol{x}, \boldsymbol{x}') + \mathbb{1}_{l,l'} \Sigma_l(\boldsymbol{x}, \boldsymbol{x}')$$

where $l' = 0,1,2,...,k$ and $\mathbb{1}_{l,l'}$ is the Kronecker delta. With the abovementioned priors, given the training points, the posterior means and variances can be computed with GP regression [37]. In this study, the hyperparameters for the multi-fidelity GP metamodel are estimated using MLE, where the mean function is assumed to be a constant and the covariance kernels are Gaussian. For more detailed description of the multi-fidelity GP metamodel, the reader is referred to Poloczek et al. [30].

For any untried point $\boldsymbol{x}^*$, the multi-fidelity GP metamodel provides the normal distribution with the posterior mean $\mu(l, \boldsymbol{x})$ and the posterior variance $\sigma^2(l, \boldsymbol{x})$. The latter is the uncertainty information of the Gaussian process metamodel which has been used in many adaptive GP reliability methods. The utilization of the uncertainty information in mfEGRA is discussed in the next section.

### 3.2 mfEGRA

Chaudhuri et al. [32] adopted the multi-fidelity GP metamodel and proposed mfEGRA. The general process of mfEGRA follows the same path as adaptive GP reliability methods described in Section 2. The main difference lies in Step (2), as for multi-fidelity GP metamodels, not only the training point but also the information source needs to be determined. The high-fidelity data are accurate but expensive; the low-fidelity data are not as accurate but less expensive. The low-fidelity data can help to explore the approximate surface of the limit state function and the high-fidelity data can help to improve the accuracy of the metamodel. Choosing the correct information source can boost the performance of the reliability analysis.

In each iteration of mfEGRA, the next 'best' training point is determined through finding the point with the maximum $EFF$ for the high-fidelity metamodel $\hat{g}(0, \boldsymbol{x})$, which is the same as in the original EGRA. Note that in the original mfEGRA, the 'best' training point is identified through solving an optimization problem with an objective function ($EFF$ in this case) to be maximized. However, in the literature of metamodel-based reliability analysis, the sampling approach, where the 'best' point is directly found from the generated candidate point population, is often used. The optimization approach used in mfEGRA can sometimes help find better points; however, the results highly rely on the performance of the optimization algorithm. And we found that in the later iterations of the refinement of the metamodel, the objective function tends to be flat and it can be difficult to find the 'best' points using the optimization approach. In addition, the optimization approach requires additional computations, while the sampling approach does not require any additional computations. Due to the efficiency and robustness, the sampling approach is used here in this study when mfEGRA is compared with the proposed method. After the 'best' point is determined, the information source is selected by measuring the KL divergence between the present metamodel GP and a hypothetical future metamodel GP when a particular information source is used to simulate the determined 'best' point. The determination of the information source is as follows.

Let $\hat{g}_P(l, \boldsymbol{x}) = \hat{g}(l, \boldsymbol{x}|\{\boldsymbol{x}^i, l^i\}_{i=1}^n)$ denote the present GP metamodel constructed with the $n$ available training points. Thus, the present high-fidelity metamodel Gaussian distribution at any point $\boldsymbol{x}$ is as follows:

$$G_P(\boldsymbol{x}) \sim N(\mu_P(0, \boldsymbol{x}), \sigma_P^2(0, \boldsymbol{x})) \tag{7}$$

where $\mu_P(0, \boldsymbol{x})$ is the posterior mean and $\sigma_P^2(0, \boldsymbol{x})$ is the posterior variance for the high-fidelity metamodel constructed with the current $n$ available training points. Assume that a hypothetical future GP metamodel is constructed with the current $n$ available training points and a possible future training point $\boldsymbol{x}^{n+1}$ from a possible future information source $l_F$. The current GP model acts as a generative model to create hypothetical future simulated data. Let $y^F$ denote the hypothetical future simulated prediction at $\boldsymbol{x}^{n+1}$ from $l_F$. It can be shown that $y^F$ follows $N(\mu_P(l_F, \boldsymbol{x}^{n+1}), \sigma_P^2(l_F, \boldsymbol{x}^{n+1}))$. Thus, a hypothetical future high-fidelity metamodel Gaussian distribution at any $\boldsymbol{x}$ is:

$$G_F(\boldsymbol{x}|\boldsymbol{x}^{n+1}, l_F, y^F) \sim N(\mu_F(0, \boldsymbol{x}|\boldsymbol{x}^{n+1}, l_F, y^F), \sigma_F^2(0, \boldsymbol{x}|\boldsymbol{x}^{n+1}, l_F, y^F)) \tag{8}$$

The posterior variance of the hypothetical future GP metamodel $\sigma_F^2(0, \boldsymbol{x}|\boldsymbol{x}^{n+1}, l_F, y^F)$ depends only on the location $\boldsymbol{x}^{n+1}$ and the source $l_F$, thus it can be replaced with $\sigma_F^2(0, \boldsymbol{x}|\boldsymbol{x}^{n+1}, l_F)$. As the posterior mean of the hypothetical future high-fidelity GP is affine with respect to $y^F$, it is a normal random variable as follows:

$$\mu_F(0, \boldsymbol{x}|\boldsymbol{x}^{n+1}, l_F, y^F) \sim N(\mu_P(0, \boldsymbol{x}), \bar{\sigma}^2(\boldsymbol{x}|\boldsymbol{x}^{n+1}, l_F)) \tag{9}$$

where $\bar{\sigma}^2(\boldsymbol{x}|\boldsymbol{x}^{n+1}, l_F)) = (\Sigma_P((0, \boldsymbol{x}), (l_F, \boldsymbol{x}^{n+1})))^2 / (\Sigma_P((l_F, \boldsymbol{x}^{n+1}), (l_F, \boldsymbol{x}^{n+1})))$ [30], [32]. Note that no new evaluations of the information are needed to construct the hypothetical future GP. The total lookahead information gain for any $\boldsymbol{x}^{n+1}, l_F$ can then be calculated by taking the expectation of the total KL divergence between $G_P$ and $G_F$, which is given by:

$$D_{IG}(\boldsymbol{x}^{n+1}, l_F) = \int_\Omega D(\boldsymbol{x}|\boldsymbol{x}^{n+1}, l_F) d\boldsymbol{x} \tag{10}$$

where $\Omega$ is the entire domain for $\boldsymbol{x}$, and $D(\boldsymbol{x}|\boldsymbol{x}^{n+1}, l_F)$ is given by:

$$D(\boldsymbol{x}|\boldsymbol{x}^{n+1}, l_F) = \log\left(\frac{\sigma_F(0, \boldsymbol{x}|\boldsymbol{x}^{n+1}, l_F)}{\sigma_P(0, \boldsymbol{x})}\right) + \frac{\sigma_P^2(0, \boldsymbol{x}) + \bar{\sigma}^2(\boldsymbol{x}|\boldsymbol{x}^{n+1}, l_F)}{2\sigma_F^2(0, \boldsymbol{x}|\boldsymbol{x}^{n+1}, l_F)} - \frac{1}{2} \tag{11}$$

If a discrete set $X \subset \Omega$ is generated via Latin hypercube sampling (LHS) to estimate $D_{IG}(\boldsymbol{x}^{n+1}, l_F)$, $D(\boldsymbol{x}|\boldsymbol{x}^{n+1}, l_F)$ can be estimated as:

$$D(\boldsymbol{x}|\boldsymbol{x}^{n+1}, l_F) = \int_\Omega D(\boldsymbol{x}|\boldsymbol{x}^{n+1}, l_F) d\boldsymbol{x} \propto \sum_{X \subset \Omega} D(\boldsymbol{x}|\boldsymbol{x}^{n+1}, l_F) \tag{12}$$

In mfEGRA, the next 'best' information source is selected by maximizing the lookahead information gain weighted by $EFF$ and normalized by the cost of the information as follows:

$$l^{n+1} = \arg\max_{l \in \{0,1,\dots,k\}} \sum_{X \subset \Omega} \frac{1}{c_l(\boldsymbol{x})} EFF(\boldsymbol{x}) D(\boldsymbol{x}|\boldsymbol{x}^{n+1}, l_F = l) \tag{13}$$

where $c_l(x)$ is the cost of model evaluation at $x$ from the information source $l$. It should be noted that the cost of the GP metamodel construction is neglected when normalizing the learning function. The rest of mfEGRA can be similar to what has been described in Section 2. When the stopping criterion of $\max(EFF) < 10^{-10}$ is satisfied, the refinement of the metamodel is stopped and the current high-fidelity GP metamodel is used to perform the reliability analysis.

When selecting the information source, the KL divergence indicates the global change for the high-fidelity GP metamodel given a specific information source being chosen. However, this "global change" is not tantamount to the global improvement and the effect of the actual improvement for the metamodel in terms of the accuracy of reliability analysis is unknown. In addition, the process of determining the 'best' training point is independent of the information from low-fidelity information sources. One can argue that the training point being chosen only based on the high-fidelity GP model is reasonable as at the end, only the high-fidelity GP model is used for the reliability analysis. However, during the process of refinement, there can be more cost-effective training points from low-fidelity models for improving the performance of the metamodel. In light of these gaps, a new learning function that chooses the training point and the information source simultaneously is proposed and a new method based on this learning function is introduced in the next section.

## 4. The proposed adaptive multi-fidelity Gaussian process reliability analysis
This section introduces the proposed adaptive multi-fidelity Gaussian process reliability analysis (AMGPRA). Firstly, a new learning function used in the proposed method is introduced in detail. Then the procedures of AMGPRA are presented.

### 4.1 Collective learning function
In this study, a new learning function is proposed for the adaptive refinement of the multi-fidelity GP metamodel for reliability analysis. This learning function has two objectives: (1) to select the training point and information source simultaneously and (2) to refine the multi-fidelity GP metamodel in a cost-effective fashion.

Existing learning functions such as $EFF$ and $U$ are intended for single-fidelity reliability analysis problems where the focus is on the selection of training points. As such, these learning functions cannot be directly used to select the information source. In addition, when considering the impact of adding a potential training point, most existing learning functions consider only the mean value and posterior variance of that single potential training point as the input, while the common stopping criteria based on the learning functions consider the set of all candidate training points $S$, e.g., $\max_{x \in S}(EFF) < 0.001$ or $\min_{x \in S}(U) > 2$. A key gap is the characterization and consideration of the global impact of adding a potential training point, i.e., the improvement in learning function values of all other points in $S$ if a candidate training point is added to the set of training points.

As mentioned in Section 3.2, for each candidate training point from an information source $\{x^{n+1}, l_F\}$, a generative model for a hypothetical future GP metamodel can be easily constructed without new evaluations of models, that is, for each $\{x^{n+1}, l_F\}$, the hypothetical future posterior variance for any point from the high-fidelity model $\sigma_F^2(0, x|x^{n+1}, l_F)$ can be obtained. According to Eq. (8), $\mu_F(0, x|x^{n+1}, l_F, y^F)$ follows a normal distribution with the mean of $\mu_P(0, x)$. The assumption that $\mu_F(0, x|x^{n+1}, l_F, y^F)$ remains the same as $\mu_P(0, x)$ is made here to avoid high computational burden. Thus, for each candidate $\{x^{n+1}, l_F\}$, the hypothetical future posterior mean $\mu_P(0, x)$ and posterior variance $\sigma_F^2(0, x|x^{n+1}, l_F)$ at any point $x$ can be calculated for the hypothetical future high-fidelity GP metamodel. And such information can be utilized to form our proposed learning function.

In single-fidelity adaptive GP reliability methods, most learning functions are de facto functions of the posterior mean $\mu_P(0, x^*)$ and posterior variance $\sigma_P^2(0, x^*)$ for any untried point $x^*$. Note that 0 indicates the high-fidelity model, which is the only model in single-fidelity problems. For instance, $EFF(x^*)$ in Eq. (3) can be written as $EFF(\mu_P(0, x^*), \sigma_P^2(0, x^*))$. Thus, in a multi-fidelity problem, for the hypothetical future high-fidelity GP given a candidate $\{x^{n+1}, l_F\}$, the $EFF$ value is $EFF(\mu_P(0, x^*), \sigma_F^2(0, x^*|x^{n+1}, l_F))$ for any untried point $x^*$. Assume a candidate $\{x^{n+1}, l_F\}$ is chosen. Adding $\{x^{n+1}, l_F\}$ can change $\sigma_P^2(0, x^*)$ to $\sigma_F^2(0, x^*|x^{n+1}, l_F)$ for all the untried points in $S$. Thus, the $EFF$ values $EFF(\mu_P(0, x^*), \sigma_F^2(0, x^*|x^{n+1}, l_F))$ will be different (or smaller) than $EFF(\mu_P(0, x^*), \sigma_P^2(0, x^*))$ for all the points in $S$. If we accumulate all the changes from all points given a $\{x^{n+1}, l_F\}$, the improvement of learning function values of all other points in $S$ when adding $\{x^{n+1}, l_F\}$ can be quantified.

Assume a learning function denoted as $lf$ is available and it had similar functioning to most learning functions for single-fidelity problems, meaning that it is a function of the posterior mean $\mu_P(0, x)$ and posterior variance $\sigma_P^2(0, x)$ (property $a$). Further assume that a larger $lf$ indicates a 'better' training point and $lf$ for an existing training point should be zero or a very small value if the noise term in the covariance matrix is considered (property *b*). A collective learning function for a candidate $\{x^{n+1}, l_F\}$ can be constructed as follows:

$$CLF(x^{n+1}, l_F) = \frac{\sum_{i=1}^{N_{mcs}} \left[ lf\left(\mu_P(0, x^{(i)}), \sigma_P^2(0, x^{(i)})\right) - lf\left(\mu_P(0, x^{(i)}), \sigma_F^2(0, x^{(i)}|x^{n+1}, l_F)\right) \right]}{N_{mcs}} \quad (14)$$

The equation measures the average improvement of learning function values of all other candidate training points in $S$ when $\{x^{n+1}, l_F\}$ is added to the training points. In other words, this collective learning function considers the global impact instead of the local impact. In practice, it is not necessary to consider all the candidate training points in $S$. The points with large $lf$ values are more 'important' than other points in terms of improving the accuracy of the metamodel. Thus, a group of $N_c$ points selected based on $lf$ can be considered here. Subsequently, the collective learning function can be given as:

$$CLF(x^{n+1}, l_F) = \frac{\sum_{i=1}^{N_c} \left[ lf\left(\mu_P(0, x^{(i)}), \sigma_P^2(0, x^{(i)})\right) - lf\left(\mu_P(0, x^{(i)}), \sigma_F^2(0, x^{(i)}|x^{n+1}, l_F)\right) \right]}{N_c} \quad (15)$$

In this study, $N_c$ is taken as 1000, which is sufficient based on authors' experience for multiple computational experiments. Note that selecting a training point from the high-fidelity information source is always more beneficial than selecting this point from a low-fidelity information source, as a high-fidelity training point would have a more direct impact on the high-fidelity GP metamodel than a low-fidelity counterpart. However, a low-fidelity training point is often significantly cheaper. The proposed collective learning function is normalized by the cost of the selected information source as given by:

$$CLF(x^{n+1}, l_F) = \frac{\sum_{i=1}^{N_c} \left[ lf\left(\mu_P(0, x^{(i)}), \sigma_P^2(0, x^{(i)})\right) - lf\left(\mu_P(0, x^{(i)}), \sigma_F^2(0, x^{(i)}|x^{n+1}, l_F)\right) \right]}{N_c c_l(x^{n+1})} \quad (16)$$

It is obvious that $EFF$ satisfies the profile described for $lf$ (property *a* and *b*). In theory, all the learning functions that satisfy the profile can be used in the collective learning function ($CLF$). To demonstrate the performance of the $CLF$, a modified version of $U$ learning function $U_m$ is designed here. Similar to other existing learning functions, $U_m$ considers training points that are close to the limit state and have large uncertainty. The mathematical expression of $U_m$ is given by:

$$U_m\left(\mu_P(0, x^*), \sigma_P^2(0, x^*)\right) = \frac{\sigma_P(0, x^*)}{\exp\left(|\mu_P(0, x^*)|\right)} \quad (17)$$

The modification is made to satisfy the profile of $lf$. Similar to $U$ learning function, $U_m$ assigns more weight to points close to the limit state with high posterior variances like $EFF$. This model for $U_m$ along with $EFF$ will be used later in the proposed $CLF$ in the computational experiments of this study.

As the $CLF$ considers the global impact of adding a training point from an information source, it should help construct the metamodel more efficiently than the approach of mfEGRA where the training point and information source are identified separately. In addition, due to the flexibility of $CLF$, it can be easily extended with any $lf$ that fits the profile to adapt to different problems. The method based on $CLF$ named AMGPRA is presented in the next subsection.

### 4.2 Application process of the proposed method
AMGPRA uses the proposed $CLF$ to adaptively refine the metamodel for reliability analysis. The general procedure is similar to the single-fidelity adaptive GP reliability methods, however, in each iteration, the information source, in addition to the training point, should be identified. The flowchart of the approach is shown in Fig. 2. For a reliability problem with $k$ fidelities, the analysis steps are summarized below:
- **Step 1**: *Generating initial candidate design samples.* First, $N_{MCS}$ candidate design samples are generated using Latin Hypercube Sampling and the set of samples is denoted as $S$.
- **Step 2**: *Selecting initial training points.* Randomly select $N_i$ points from $S$. $N_i$ is determined as min $\{12, (D + 1)(D + 2)/2\}$, where $D$ is the number of random variables. Evaluate the $N_i$ points for models for all $k$ fidelities to get the initial training set $\{x^i, l_i\}_{i=1}^{kN_i}$.
- **Step 3**: *Constructing multi-fidelity GP metamodel.* Construct the multi-fidelity GP metamodel $\hat{g}(l, x|\{x^i, l^i\}_{i=1}^{kN_i})$ with current $X_{tr}$.
- **Step 4:** *Multi-fidelity GP metamodel prediction.* The responses $\mu_P(0, x)$ and posterior variance $\sigma_P^2(0, x)$ are obtained from the current high-fidelity GP metamodel $\hat{g}(0, x|\{x^i, l^i\}_{i=1}^{kN_i})$ for every point in $S$. According to responses $\mu_P(0, x)$, the failure probability $\hat{P}_f$ is estimated via MCS.
- **Step 5:** *Checking the stopping criterion based on the maximum error.* Check the stopping criterion:

$$\max(EFF) < 0.001 \tag{18}$$

If not satisfied, go to Step 7, otherwise, go to Step 6.
- **Step 6**: *Checking the coefficient of variation of the failure probability.* The sufficiency of the population of $S$ is checked using:

$$COV_{\hat{P}_f} = \sqrt{\frac{1 - \hat{P}_f}{N_{MCS}\hat{P}_f}} \tag{19}$$

If $COV_{\hat{P}_f}$ is smaller than the predefined threshold 5%, then go to Step 9. Otherwise, an additional number $N_{\Delta_S}$ of candidate design samples $\Delta_S$ should be added to $S$, and the process should move back to Step 4.
- **Step 7**: *Identification of the next training point and information source.* Identify $N_c$ points with the largest $lf$ values in $S$ and apply the proposed $CLF$ to these points and select the next training point $x^{n+1}$ from the information source $l_F$ with the largest $LF$ value. Add $\{x^{n+1}, l_F\}$ to the set of training points. Note that when the high-fidelity model is chosen as the training source, add the training point from all the fidelities to the training point set, as the costs of low-fidelity training points are often significantly lower than the high-fidelity ones and these low-fidelity training points can help construct a more accurate model discrepancy for the multi-fidelity metamodel.
- **Step 8**: *Update the set of training points:* Update the set of training points with the identified point and information source. Go back to Step 3.
- **Step 9**: *End.* Report $\hat{P}_f$

The proposed method selects the training point and the information source simultaneously when considering the global impact of adding the training point with the help of the proposed $CLF$. The performance of the proposed method is demonstrated through four computational experiments. The details are presented in the next section.

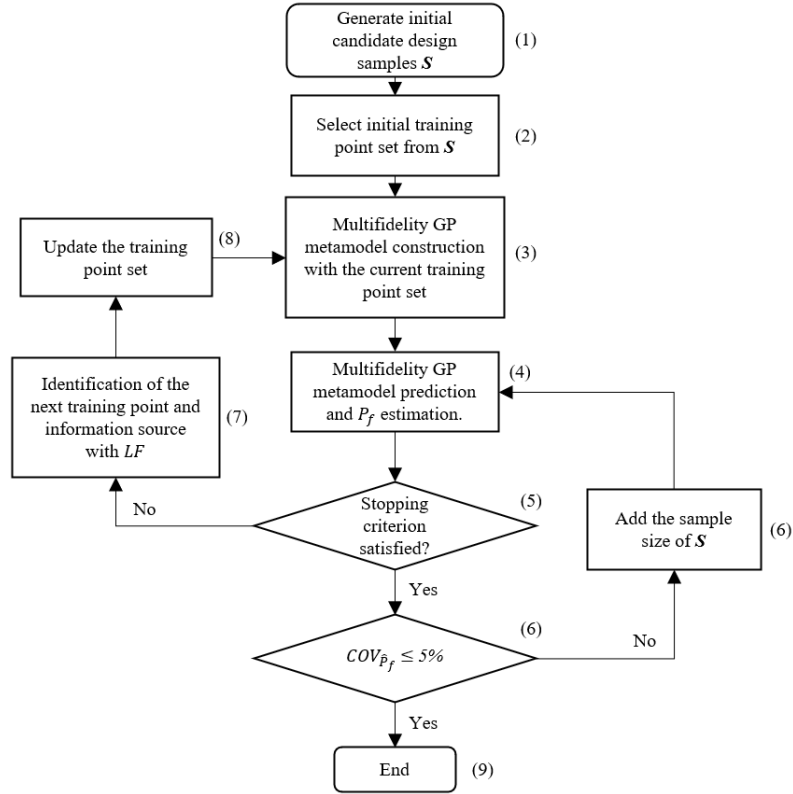

Fig.2 Flowchart of proposed approach

## 5. Computational experiments
In order to test the performance of the proposed method, four computational experiments are investigated. Results are mainly compared with one of the most popular single-fidelity GP reliability methods: AK-MCS-$EFF$ and the state-of-the-art multi-fidelity GP reliability method mfEGRA. For fair comparison with mfEGRA, the same adaptive multi-fidelity GP model and initial training points are shared between the two methods. In addition, the same stopping criterion ($\max(EFF) < 0.001$) is used for both methods. Note that the cost $c\_l$ only includes the model evaluation and the cost of constructing the multi-fidelity GP model is not included, as it is negligible compared to the cost of model evaluations. The computations are performed using MATLAB 2019b. All computations are repeated for 20 times in this study.

### 5.1. A multimodal function example
The first example investigates the performance of the proposed method for a highly nonlinear and multimodal function. The problem appeared originally in Bichon et al. [12]. The original limit state is given by:

$$g_0(x) = 2 - \frac{(x_1^2 + 4)(x_2 - 1)}{20} + \sin\left(\frac{5x_1}{2}\right) \tag{20}$$

where the distribution of $x_1$ is normal with the mean of 1.5 and standard deviation of 1 and $x_2$ is also normal with the mean of 2.5 and standard deviation of 1. The original function is regarded as the high-fidelity model herein. Chaudhuri et al. [32] added two low-fidelity models as follows:

$$\begin{aligned} g_1(x) &= g_0(x) - \sin\left(\frac{5x_1}{22} + \frac{5x_2}{44} + \frac{5}{4}\right) \\ g_2(x) &= g_0(x) - \sin\left(\frac{5x_1}{11} + \frac{5x_2}{22} + \frac{35}{11}\right) \end{aligned} \tag{21}$$

The cost of each model is assumed to remain constant over the entire domain. The costs are $c_0 = 1$, $c_1 = 0.1$, and $c_2 = 0.01$. Fig. 3 shows the response of the three models over the $x_1$-$x_2$ domain. For this example, the number of initial training points is set to 6 for all methods, and the initial number of candidate design points is $N_{MCS} = 10^4$ with $N_{\Delta_S} = 10^4$. The results obtained from MCS with a sample size of $10^6$ are used as the reference.

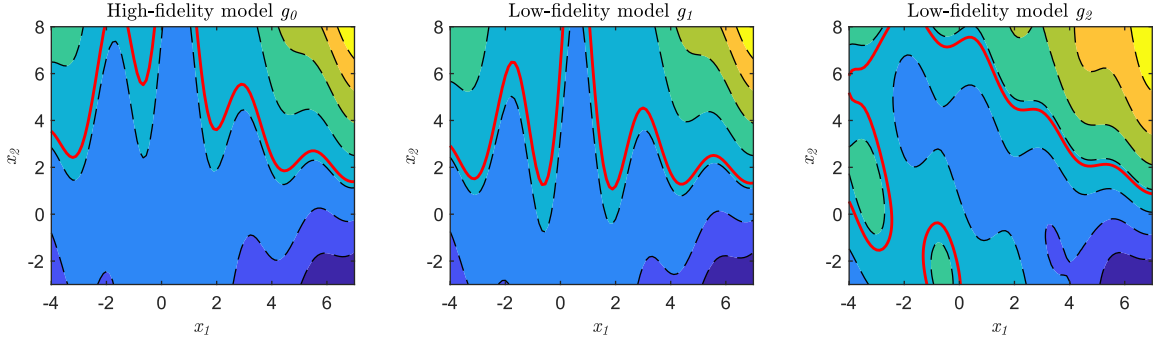

Fig. 3 Contours of models with different fidelities. The red lines indicate the limit states.

Both $EFF$ and $U_m$ are used as $lf$ in $CLF$. Hereafter, the proposed methods with $CLF$ using $EFF$ and $U_m$ are denoted as AMGPRA-$LF$-$EFF$ and AMGPRA-$LF$-$U_m$, respectively. In this example, the high-fidelity model $g_0$ and low-fidelity model $g_1$ are considered first to construct a bi-fidelity reliability problem. The results of the proposed methods and other methods including AMK-MCS-$EFF$ by Yi et al. [33] are shown in Table 1. In this case, AMGPRA-$LF$-$EFF$ performs slightly better than AMGPRA-$LF$-$U_m$, and both proposed methods outperform the single-fidelity methods and other multi-fidelity methods. Both AMGPRA methods cost less than all other methods. In addition, AMGPRA-$LF$-$EFF$ provides the lowest average relative error among all multi-fidelity methods.

Table 1 Results of the multimodal function for different methods

| Methods | Costs | $\hat{P}_f$ ($\times 10^{-2}$) | Average relative error (%) |
|---|---|---|---|
| MCS | $10^6$ | 3.13 | - |
| AK-MCS-$EFF$ | 45.2 | 3.13 | 0.04 |
| AMK-MCS-$EFF$ | 14.31 (9.33 + 49.77 × 0.1) | 3.13 | 0.64 |
| mfEGRA | 13.31 (10.1 + 30.1 × 0.1) | 3.16 | 0.06 |
| AMGPRA-$LF$-$EFF$ | 12.58 (9.9 + 26.8 × 0.1) | 3.14 | 0.03 |
| AMGPRA-$LF$-$U_m$ | 12.86 (9.9 + 29.55 × 0.1) | 3.12 | 0.06 |

When considering all three models $g_0$, $g_1$ and $g_2$, the results of the proposed methods along with other methods are shown in Table 2. It can be seen that with the lowest fidelity model $g_2$ introduced, the costs decrease compared to the bi-fidelity problem for all methods. The proposed AMGPRA methods still cost less than mfEGRA, although the difference is not as significant as the bi-fidelity case. This can be due to the fact that all methods are approaching the 'optimal' cost when there are three fidelities present. Still, it can be observed that AMGPRA methods tend to use more lowest fidelity training points to explore the domain than mfEGRA. The ratios of training points for $g_2$ to those for $g_1$ are 2.47 and 2.42 for AMGPRA-$LF$-$EFF$ and AMGPRA-$LF$-$U_m$, respectively, while this ratio is only 1.63 for mfEGRA. Fig. 4 shows the progress of AMGPRA-$LF$-$EFF$ at several iterations. In this case, AMGPRA-$LF$-$EFF$ uses a total of 9 evaluations of $g_0$, 20 evaluations of $g_1$ and 46 evaluations of $g_2$ to reach a value of $EFF$ below $10^{-3}$ and a relative error of zero.

Table 2 Results of the multimodal function for different methods

| Methods | Costs | $\hat{P}_f$ ($\times 10^{-2}$) | Average relative error (%) |
|---|---|---|---|
| mfEGRA | 12.87 (10.2 + 22.1 × 0.1 + 36 × 0.01) | 3.11 | 0.02 |
| AMGPRA-$LF$-$EFF$ | 12.32 (10 + 18.6 × 0.1 + 46 × 0.01) | 3.10 | 0.02 |

| AMGPRA-$LF$-$U_m$ | 12.41 (10.05+19 × 0.1 + 46 × 0.01) | 3.12 | 0.03 |

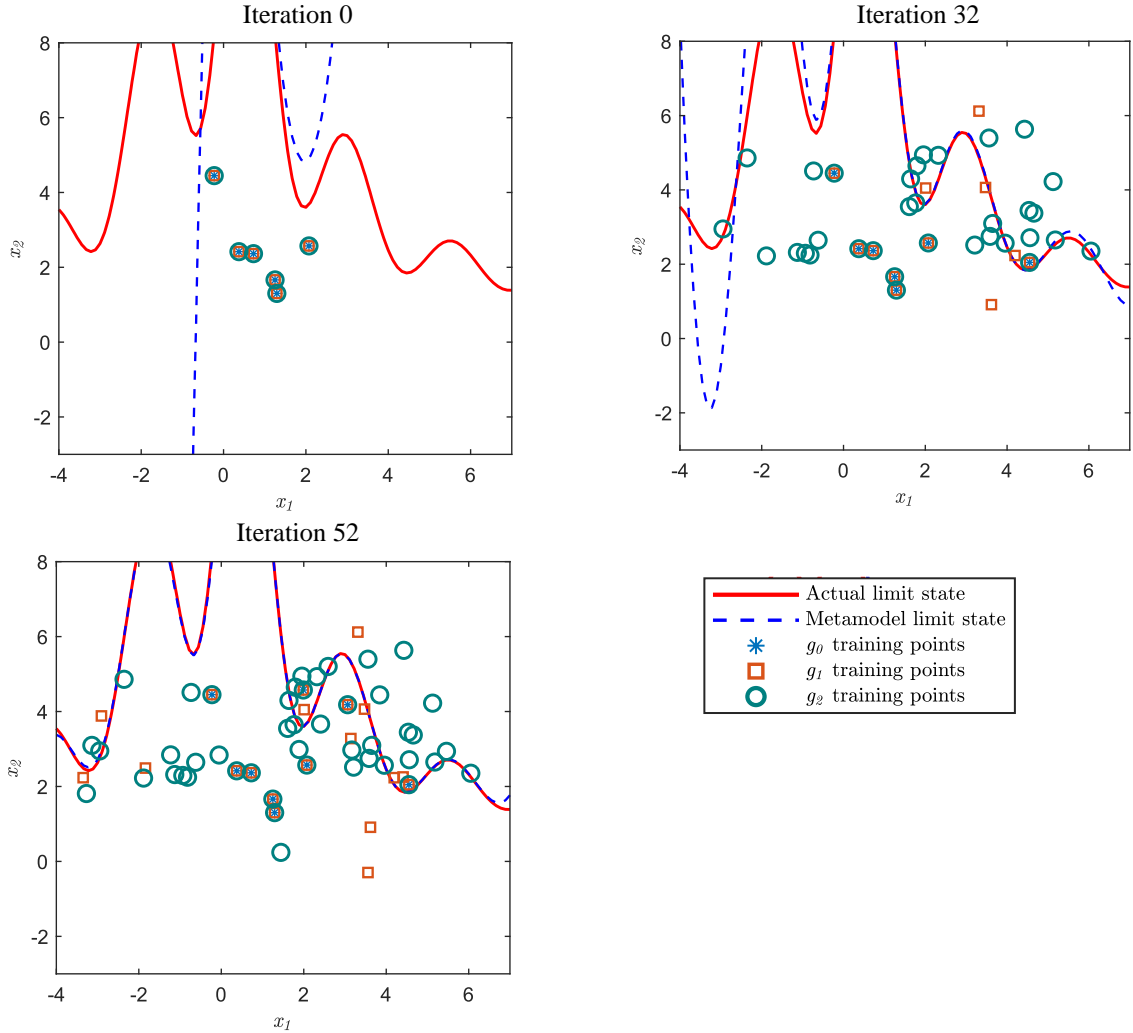

Fig. 4 Progress of AMGPRA-$LF$-$EFF$ at several iterations

### 5.2. An oscillator example

The second example investigates the performance of the proposed method for an undamped single degree of freedom system (Fig. 5) with six random variables. The example is modified to have a lower failure probability from the original example in [10], [38]–[41]. The limit state function for the high-fidelity model is described below:

$$g_0(k_1, k_2, m, r, t_1, F_1) = 3r - \left| \frac{2F_1}{m\omega_0^2} \sin\left(\frac{\omega_0 t_1}{2}\right) \right| \tag{22}$$

where $\omega_0$ is the system frequency, which is determined using $\sqrt{\frac{k_1+k_2}{m}}$. All random variables in this example follow normal distributions. The properties of the six random variables are summarized in Table 3. In this example, two low-fidelity models are provided as follows:

$$\begin{aligned} g_1(k_1, k_2, m, r, t_1, F_1) &= g_0(k_1, k_2, m, r, t_1, F_1) - \frac{1}{15}\sin\left(\frac{\omega_0 t_1}{2}\right) \\ g_2(k_1, k_2, m, r, t_1, F_1) &= g_0(k_1, k_2, m, r, t_1, F_1) - \frac{2}{15}\sin\left(\frac{\omega_0 t_1}{2}\right) \end{aligned} \tag{23}$$

The cost of each model is assumed to remain constant over the entire domain. These costs are $c_0 = 1, c_1 = 0.1$, and $c_2 = 0.01$. For this example, the number of initial training points is set to 12 for all methods, and the initial number of candidate design points is $N_{MCS} = 10^4$ with $N_{\Delta_S} = 10^4$.

For multi-fidelity methods, all three models are used. The results of the proposed methods along with other methods are shown in Table 4. It can be seen that without adding new high-fidelity training points, all multi-fidelity methods achieve satisfactory results with only adding low-fidelity training points. Thus, there are not significant cost differences between the proposed methods and mfEGRA. However, with a similar cost, the proposed method achieves lower relative errors compared to mfEGRA. It can be observed that the multi-fidelity methods perform significantly better than the single-fidelity methods.

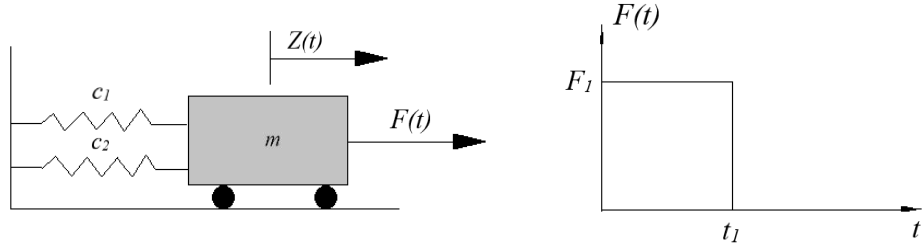

Fig. 5 Oscillator

Table 3 Random variables in the oscillator

| Random variable | Distribution | Mean | Standard deviation |
|---|---|---|---|
| $m$ | Normal | 1 | 0.05 |
| $k_1$ | Normal | 1 | 0.1 |
| $k_2$ | Normal | 0.1 | 0.01 |
| $r$ | Normal | 0.65 | 0.05 |
| $F_1$ | Normal | 1 | 0.2 |
| $t_1$ | Normal | 1 | 0.2 |

Considering the observation that no additional high-fidelity training points are added when 12 initial high-fidelity training points are used in our experiments, it may not be necessary to have 12 initial high-fidelity training points. To better examine the performance of the proposed methods relative to mfEGRA, the initial number of training points for each model is reduced from 12 to 8. The results are shown in Table 4. It can be seen that all the multi-fidelity methods outperform the single-fidelity method by a large margin in terms of cost. AMGPRA-$LF$-$mU$ has the lowest cost among the multi-fidelity methods, and the errors of the three multi-fidelity methods are close to each other. The boxplots regarding the costs and errors are also presented in Fig. 6. It can also be observed that AMGPRA-$LF$-$EFF$ provides the most stable results in terms of both cost and error.

Table 4 Results of the oscillator for different methods

| Methods | Costs | $\hat{P}_f$ ($\times 10^{-4}$) | Average relative error (%) |
|---|---|---|---|
| MCS | $10^6$ | 8.2 | - |
| AK-MCS-$EFF$ | 50.8 | 8.2 | 0.25 |
| mfEGRA | 10.14 (8.6+9.9 × 0.1 + 52.6 × 0.01) | 8.2 | 0.51 |
| AMGPRA-$LF$-$EFF$ | 9.49 (8+9 × 0.1 + 58.8 × 0.01) | 8.2 | 0.54 |
| AMGPRA-$LF$-$mU$ | 10.23 (8.3+11 × 0.1 + 81.9 × 0.01) | 8.2 | 0.48 |

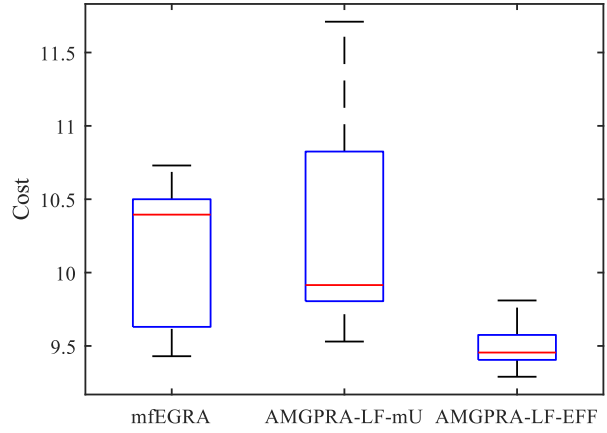

(a) Boxplot for the costs of different methods

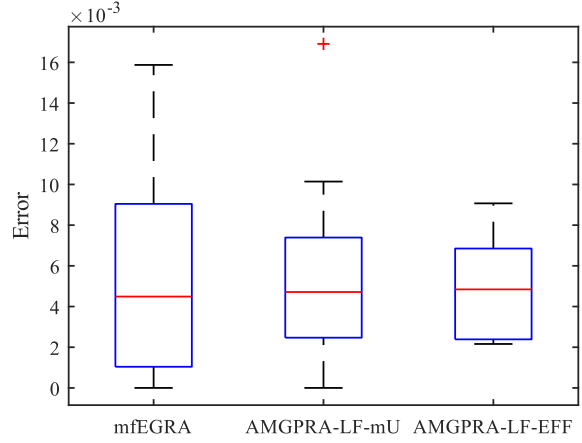

(b) Boxplot for the errors of different methods

Fig. 6　Boxplots for different methods

### 5.3. A ten-dimensional problem

The original problem in example three was presented in [10], [42], [43] with the following form:

$$g_0(x) = (n + 3\sigma\sqrt{n}) - \sum_{i=1}^{n} x_i \tag{24}$$

where the random variables, $x_i$s follow lognormal distribution with the mean of 1 and standard deviation $\sigma = 0.2$, and $n$ is the number of random variables. Here, $n$ is set to be 10. In this example, one low-fidelity model is provided as follows:

$$g_1(x) = (n + 3\sigma\sqrt{n}) - a\sum_{i=1}^{n} x_i \tag{25}$$

where $a$ is the accuracy parameter and is set to 0.9 here. The cost of each fidelity model is assumed to remain constant over the entire domain and is given by $c_0 = 1$ and $c_1 = 0.05$. For this example, the number of initial training points is set to 12 for all methods, and the initial number of candidate design points is $N_{MCS} = 10^5$ with $N_{\Delta_S} = 10^5$.

Results of the proposed methods along with other methods are shown in Table 5. Both proposed methods are more efficient with lower costs. In this case, the ratios of the number of training points from $g_1$ to those for $g_0$ are 2.4 and 2.35 for AMGPRA-$LF$-$EFF$ and AMGPRA-$LF$-$U_m$, respectively, while this ratio is 1.78 for mfEGRA, which is lower than the former two. Taking more advantage of low-fidelity training points, the proposed methods achieve lower errors while incurring lower total costs.

Table 5 Results of the ten-dimension problem for different methods

| Methods | Costs | $\hat{P}_f$ ($\times 10^{-3}$) | Average relative error (%) |
|---|---|---|---|
| MCS | $10^6$ | 2.73 | - |
| AK-MCS-$EFF$ | 62.45 | 2.75 | 0 |
| mfEGRA | 18.16 (16.7 + 29.2 × 0.05) | 2.78 | 0.02 |
| AMGPRA-$LF$-$EFF$ | 16.8 (15 + 36 × 0.05) | 2.74 | 0 |
| AMGPRA-$LF$-$U_m$ | 16.76 (15 + 35.2 × 0.05) | 2.70 | 0.01 |

To investigate the impact of the cost-ratio of high-fidelity and low-fidelity models on the results of the proposed method, a parametric study for AMGPRA-$LF$-$EFF$ is performed for this example. The cost of the low-fidelity model $c_1$ is set to vary from 0.04 to 0.5 resulting in the cost-ratio ranging from 2 to 25. The experiment is performed with 10 repeated trials to obtain the average results. The proposed method is able to provide the failure probability estimates with the error of 0.0% for all cost ratios. However, the costs are different with different cost ratios. The results are visualized in Fig. 7. It can be observed that when the cost of the low-fidelity model is lower (i.e., the cost ratio is larger), the proposed method tends to yield the result with a lower cost as shown in Fig. 7 (a). For such cases, the proposed method tends to use more low-fidelity training points as shown in Fig. 7 (b). As the cost of the low-fidelity model increases, the proposed method selects a higher percentage of training points from the high-fidelity model.

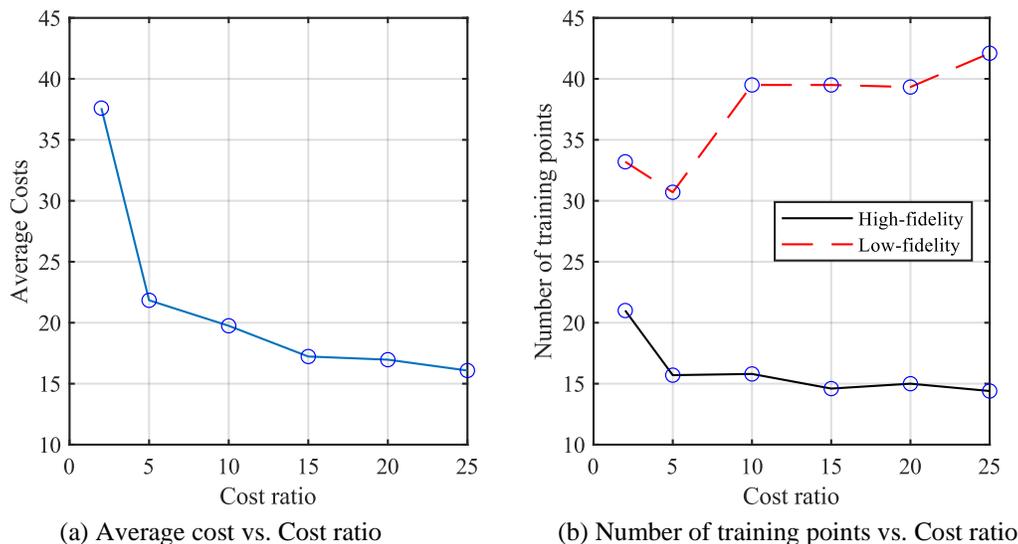

(a) Average cost vs. Cost ratio      (b) Number of training points vs. Cost ratio

Fig. 7 Trends for the impact of different cost ratios of high-fidelity to low-fidelity models on the overall performance of multi-fidelity reliability analysis

To further investigate the influence of the accuracy of the low-fidelity model, another parametric study is performed for this example. The accuracy constant $a$ in the low-fidelity model is set to vary from 0.5 to 0.9 with 0.5 making the low-fidelity model the least accurate and 0.9 making the low-fidelity model the most accurate, respectively. The results are visualized in Fig. 8. As can be observed in Fig. 8 (a), when the accuracy is increased, the proposed method tends to have a lower cost. In Fig. 8 (b), the number of training points from the high-fidelity model gradually decreases as the accuracy of the low-fidelity model increases.

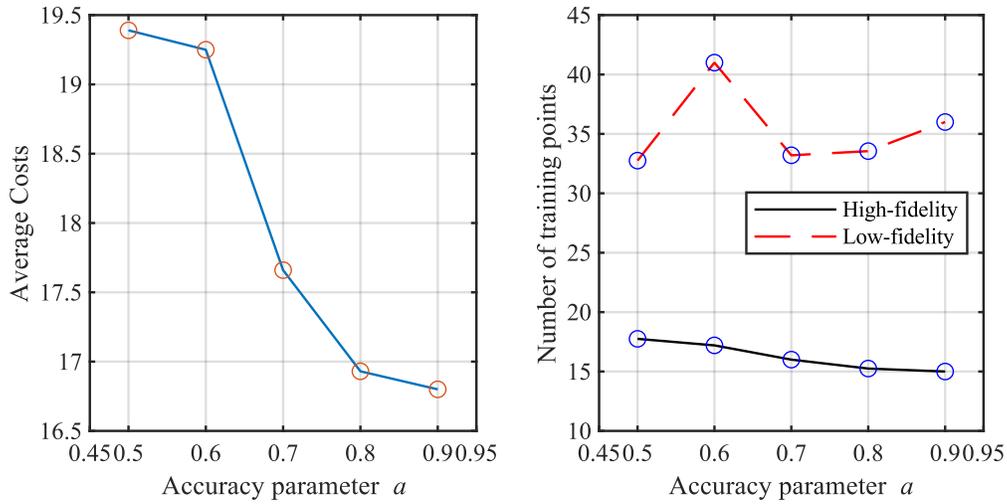

(a) Average cost vs. Accuracy  (b) Number of training points vs. Accuracy

Fig. 8  Trends for the impact of different levels of accuracy of the low-fidelity model on the overall performance of multi-fidelity reliability analysis

### 5.4. A transmission tower problem

The last example is a practical engineering problem for a transmission tower. Electric power systems are key infrastructures that support many critical activities and operations in various sectors of the society. Failure of transmission towers can result in considerable economic loss and societal and economic disruptions. Although transmission towers are often designed to withstand strong wind hazards, they have experienced damage and failures during extreme events such as hurricanes, tornados and downbursts [44]–[47]. The estimation of the reliability of transmission towers through probabilistic models is needed for risk and resilience analysis and decision-making purposes. Such reliability analyses need to be performed for many configurations of towers in the power grid, thus, efficient methods that can accurately estimate the failure probability are desired. The finite element (FE) models used in the reliability analyses are key to obtain accurate estimations. Darestani et al. [48] proposed an approach to modeling transmission towers considering involved complexities, including buckling, joint slippage, and joint failure, and their effects on the hurricane performance of towers. The objective of this computational experiment is to perform multi-fidelity reliability analysis using the high-fidelity FE model introduced in Darestani et al. [48] with assistance from a lower-fidelity FE model.

The double-circuit vertical lattice tower considered in this study is shown in Fig. 9. The 27.43-meter tower carries six lines of Drake conductors and two lines of neutral optical ground wire conductors with a span length of 258 meters. More details about the tower can be found in Darestani et al. [48]. The FE models for the transmission tower are constructed in OpenSEES platform [49]. The *BUCL&SLIPF* FE model in [48] is a high-fidelity computational model that captures buckling, joint slippage, and joint failure. On the other hand, *NBUCL&NSLIP* in [48] is a low-fidelity FE model that does not consider buckling effects, joint slippage, and joint failure. *BUCL&SLIPF* and *NBUCL&NSLIP* are considered high-fidelity and low-fidelity models respectively for multi-fidelity reliability analysis. For details of the two models, the reader is referred to Darestani et al. [48].

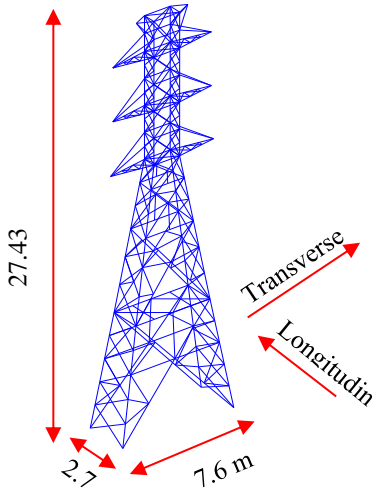

Fig. 9　Double-circuit vertical lattice tower.

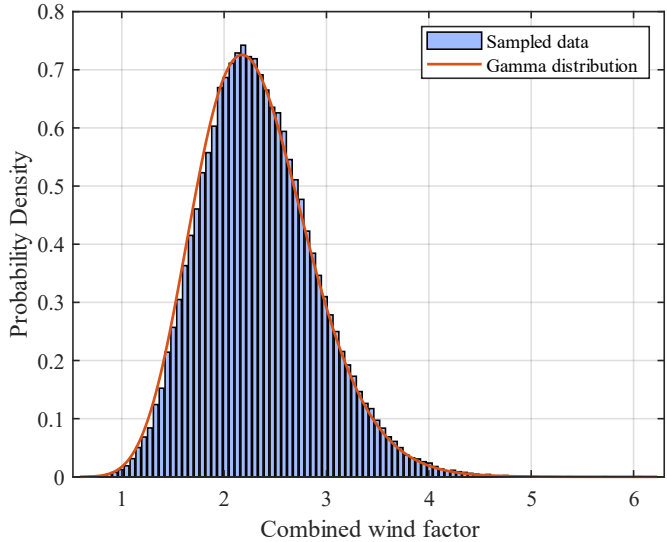

Fig. 10　Fitted distribution for the combined wind factor.

　　The objective of this example is to investigate the probability of failure of the tower under transverse winds with the design wind speed of 120 mph. The random variables considered here for the tower are modulus of elasticity $E$, yield stress of the main legs $f_{ym}$ and yield stress of other elements $f_{yb}$. The uncertainty in the wind-induced loading is also considered here. The multiplicative factors in Ellingwood and Tekie [50], including gust effect factor $G$, force coefficient $C_f$, velocity pressure exposure coefficient $K_z$ and wind directionality factor $K_d$, are combined into a single wind factor ($F_w$) to reduce the dimension of the analysis [51]. This factor is found to follow a gamma distribution with a mean of 2.31 and a coefficient of variation of 0.244. These parameters are determined using 10,000 realizations of the product of the four randomly generated factors. The fitted pdf can be found in Fig. 10. As pile foundations in stiff soil provide close to rigid foundation performance for wind hazards, it is assumed that the foundation of the tower is rigid and thus the soil and foundation are not modeled. The statistical characteristics of the four random variables can be found in Table 6. The limit state function considered here is given by:

$$g(\pmb{x}) = F_L - 1 \tag{26}$$

where $F_L$ is the load factor corresponding to the top node displacement of 0.25 m in the pushover analysis. This predefined displacement criterion corresponds to the serviceability failure of the tower, where some elements of the tower have sustained inelastic deformations. The tower in this state requires repair actions to ensure the safety of the tower against future hazards. The costs of evaluations of the limit state functions for the models of the two fidelities are considered based on the runtime of the analyses. The runtime of the high-fidelity model is approximately ten times the runtime of the low-fidelity model, thus, the cost of the high-fidelity model is set as 1 and the one for the low-fidelity model is set as 0.1. For this example, the number of initial training points is set to 12 for all methods, and the initial number of candidate design points is $N_{MCS} = 10^4$ with $N_{\Delta_S} = 10^4$.

Considering that AMGPRA-$LF$-$EFF$ outperformed AMGPRA-$LF$-$U_m$ in most previous cases, only AMGPRA-$LF$-$EFF$ is used here to compare with mfEGRA. MCS is not available as the reference due to the prohibitive computational cost of running a large number of simulations. AK-MCS is not able to converge even with 300 high-fidelity training points. The failure to converge can be attributed to two factors. Firstly, the convergence issue may be a common problem for the analysis of nonlinear structures when using AK-MCS, as the smoothness of the surface of the model is affected by the nonlinearity of the structure, hence the difficulty in converging for the GP model. Secondly, the noise in the results of the finite element analysis may generate an unsmooth surface, leading to the difficulty of convergence. Even when the noise term is added to the GP metamodel with UQLab [52], AK-MCS is still not able to converge. AMGPRA-$LF$-$EFF$ is the only method that is able to converge in this experiment, as the multi-fidelity GP model used here has a noise term in the covariance matrix, avoiding the overfitting that can hinder the convergence. On the other hand, mfEGRA does not function properly in this experiment. The reason is that mfEGRA separates the point and information selection and only checks each potential training point individually. When the point selected is considerably close to the existing training point, the posterior variance of the hypothetical future GP metamodel can be extremely close to the one of the current model, resulting in the situation where $D(x|x^{n+1}, l_F)$ in Eq. (12) is zero for both high-fidelity and low-fidelity models. Thus, the selection of information source cannot be performed here. AMGPRA-$LF$-$EFF$ is able to avoid such a situation as it simultaneously selects the training point and information source and $CLF$ considers the global impact instead of the local impact of an individual point.

The results of the reliability analysis are presented in Table 7. It can be observed that AMGPRA-$LF$-$EFF$ is able to provide an estimate of the failure probability with only 18.1 high-fidelity training points and 25.6 low-fidelity training points on average.

Table 6 Random variables in transmission tower problem

| Random variable | Distribution | Mean | Standard deviation | Reference |
| --- | --- | --- | --- | --- |
| $E$ | Lognormal | $2 \times 10^{11}$ ($N/m^2$) | $1.2 \times 10^9$ ($N/m^2$) | |
| $f_{ym}$ | Lognormal | $4.02 \times 10^8$ ($N/m^2$) | $4.02 \times 10^7$ ($N/m^2$) | ASCE 07 [53] and Wong and Miller [54] |
| $f_{yb}$ | Lognormal | $2.9 \times 10^8$ ($N/m^2$) | $2.9 \times 10^7$ ($N/m^2$) | |
| $F_w$ | Gamma | 3.22 | 0.65 | |

Table 7 Results of the transmission tower problem

| Methods | Costs | $\hat{P}_f$ ($\times 10^{-2}$) |
| --- | --- | --- |
| AK-MCS-$EFF$ | - | - |
| mfEGRA | - | - |
| AMGPRA-$LF$-$EFF$ | 28.5(24.7+37.9 × 0.1) | 2.46 |

## 6. Conclusions

For analyzing the likelihood of an event, e.g., a failure, models with different levels of fidelities are often available. The low-fidelity models, albeit with lower accuracy, are significantly cheaper to evaluate and can help the metamodel to explore the entire domain more efficiently. On the other hand, high-fidelity models are accurate but are expensive to evaluate. This paper proposed an adaptive multi-fidelity GP reliability method called AMGPRA to analyze the reliability of problems when multiple models with different fidelities are available. The proposed method adaptively refines a multi-fidelity GP metamodel until a stopping criterion that reflects on the accuracy of the model is satisfied. A new approach to selection of best training points called collective learning function ($CLF$) is proposed for the refinement of the metamodel. The model can use any learning function that satisfies a certain profile as the core. The proposed $CLF$ characterizes the global impact of adding training points in contrary to local impacts in prior studies, thus enabling simultaneous selection of the training point and information source. This capability leads to selection of more effective training points as compared to the sequential selection of data source and training point in existing methods. The proposed method provides a new direction for learning functions for the refinement of metamodels and can easily incorporate other learning functions. The performance of the proposed method is demonstrated through three mathematical experiments and one practical engineering problem concerning the wind reliability of a real transmission tower. Results indicated that AMGPRA achieves similar or better accuracy with lower computational costs compared to the state-of-the-art method mfEGRA. Moreover, the proposed method tends to take advantage of the low-fidelity information source more to facilitate the refinement of the surrogate model.

In this study, only $EFF$ and a modified version of $U$ learning function are used as the core of $CLF$. Alternative and potentially more suitable learning functions will be investigated as the cores of $CLF$ in future research. Moreover, the proposed method is constrained in treating high dimensional problems due to the general limitation of Gaussian process metamodels. Expanding the capabilities of multi-fidelity reliability analysis to high dimensional problems is an important future research direction.

**Acknowledgments**
This research has been partly funded by the U.S. National Science Foundation (NSF) through awards CMMI-1635569, 1762918, and 2000156. In addition, this work is supported in part by Lichtenstein endowment at The Ohio State University. These supports are greatly appreciated. In addition, the help from Dr. Yousef M. Darestani and Mr. Ashkan B. Jeddi for the finite element models of the transmission tower is greatly appreciated.